\documentclass[runningheads]{llncs}
\usepackage{graphicx}

\usepackage{tikz}
\usepackage{comment}
\usepackage{amsmath,amssymb} 
\usepackage{color}

\def\ssfp{SMOT}

\usepackage[width=122mm,left=12mm,paperwidth=146mm,height=193mm,top=12mm,paperheight=217mm]{geometry}

\begin{document}
\pagestyle{headings}
\mainmatter

\title{SMOT: Single-Shot Multi Object Tracking} 


\author{
Wei Li\and
Yuanjun Xiong \and
Shuo Yang\and
Siqi Deng\and
Wei Xia
}

\institute{{\tt\small wayl, yuanjx, shuoy, siqideng, wxia@amazon.com}\\Amazon AWS AI}

\maketitle
\begin{abstract}
We present single-shot multi-object tracker (\ssfp), a new tracking framework that converts any single-shot detector (SSD) model into an online multiple object tracker, which emphasizes simultaneously detecting and tracking of the object paths. Contrary to the existing tracking by detection approaches~\cite{Wojke2018deep,zhang2016tracking,lin2018prior} which suffer from errors made by the object detectors, \ssfp~adopts the recently proposed scheme of tracking by re-detection. We combine this scheme with SSD detectors by proposing a novel tracking anchor assignment module. With this design \ssfp  is able to generate tracklets with a constant per-frame runtime. A light-weighted linkage algorithm is then used for online tracklet linking. On three benchmarks of object tracking: Hannah, Music Videos, and MOT17, the proposed \ssfp~ achieves state-of-the-art performance. 

\end{abstract}

\section{Introduction}

Similar to the role of object detection in image-based visual analysis, object tracking is an essential component for visual understanding in videos~\cite{tapaswi2019video}. Its outputs are subsequently used in visual attribute analysis~\cite{kanade2000comprehensive,shen2017learning} and instance recognition~\cite{deng2019arcface,li2014deepreid}. 
The major goals of a multi-object tracker are: 1) to distinguish different objects into different tracks; 2) to keep the same object in a single track. 
Despite the rapid advance of generic object tracking algorithms ~\cite{bergmann2019tracking,feichtenhofer2017detect,zhang2016tracking}, designing a multi-object tracking approach to achieve these goals still remains a difficult problem.
Besides, due to the strict latency requirement of the video processing task, an multi-object tracking approach is also expected to have: 1) fast and  stable runtime speed; 2) accurate and robust localization accuracy. 3) The capability for online processing. 

\begin{figure}[t]
      \centering
      \includegraphics[width=1\linewidth]{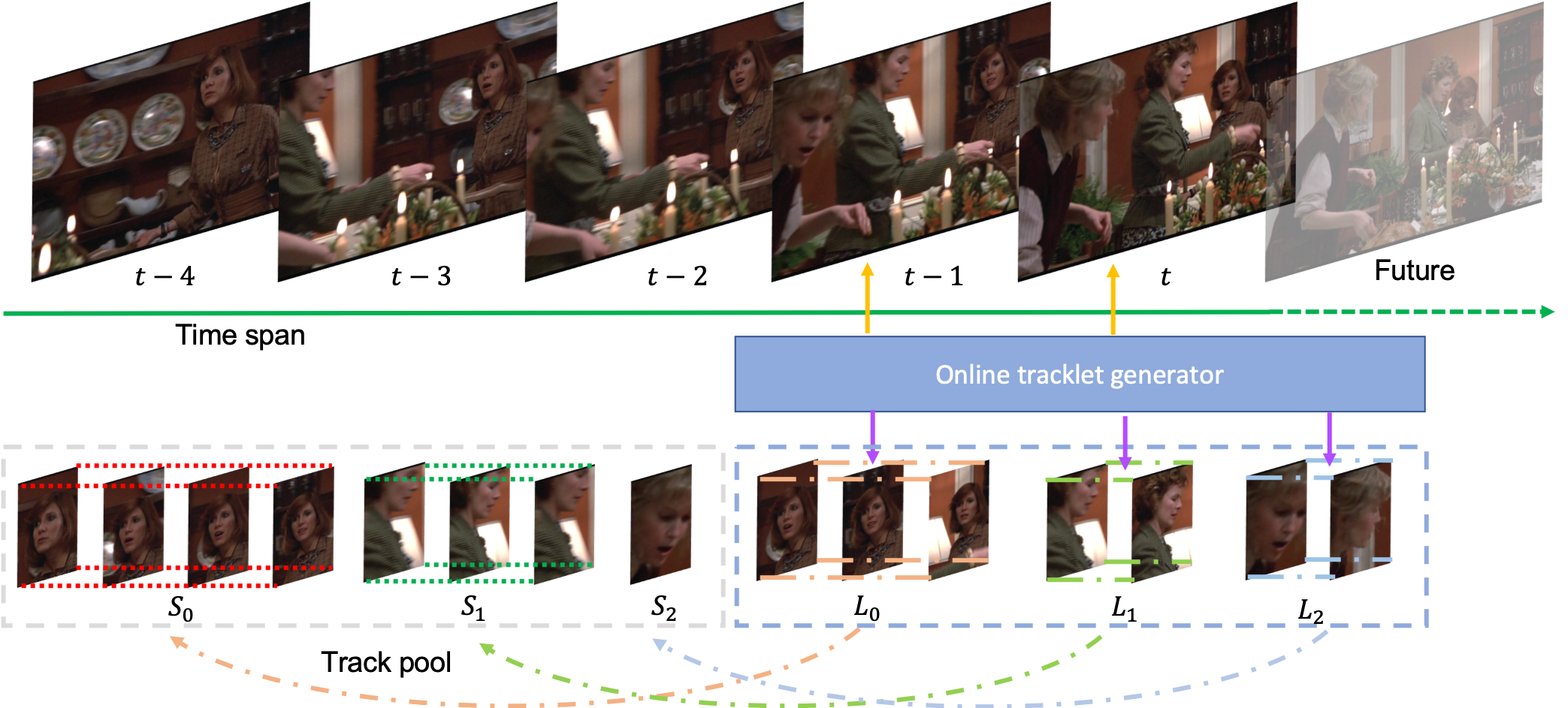}
      \centering
      \caption{The proposed \ssfp  consists of two stages. The first stage generates temporally consecutive tracklets by exploring the temporal and spatial correlations from previous frame. The second stage performs online linking of the tracklets to generate a face track for each person (better view in color).}
      \label{fig:teaser}
\end{figure}

Previous multi-object tracking algorithms~\cite{lin2018prior,zhang2016tracking,wu2013simultaneous,weng2020gnn3dmot,braso2020learning,zhang2019robust,pang2020tubetk,yin2020unified}rely on a three-stage scheme ( detection, tracklet generation and linking), also known as \emph{tracking by detection}. The first stage exercises a \emph{per-frame object detector} to localize object bounding boxes in each frame. The second stage, \emph{tracklet generation}, merges detection results to create a set of tracklets, \emph{i.e.}, short tracks, based on short-term cues. These short-term cues include motion~\cite{wojke2017simple}, appearance features~\cite{Wojke2018deep}, and geometrical affinity~\cite{sun2019deep}. Then the third stage, \emph{long-term tracklets linking}, links the tracklets in the whole video to form long-term tracking outputs. The three-stage tracking-by-detection scheme is known to provide reasonable tracking quality. However, there are several limitations in this scheme. Firstly, the division of object detection and tracklet generation makes the tracking quality highly reliant on the detector's capability in accurately detecting objects against all video-specific nuances, such as motion blurriness, large pose variations, and video compression artifacts. These detectors~\cite{ren2015faster,liu2016ssd}, however, are mostly designed for still-images and operated in a per-frame manner. Without adaptation, it is hard to utilize temporal context to overcome these nuances. Secondly, it is impossible to recover an object bounding box through temporal context once it is missed by the detector. Thirdly, low-cost matching techniques (\emph{e.g.} IoU matching in~\cite{wojke2017simple}) used in the tracklet generation process are prone to false matches, especially in occluded cases thus creating tracks with multi objects.

In this work, we propose the single-shot multi object tracker (\ssfp), a new framework for multi object tracking, to address the above issues. It is composed of two cascaded stages, the \emph{tracklet generation} stage and the \emph{tracklet linking} stage, in contrast to the three stages in previous mentioned tracking by detection scheme. We design the tracklet generation stage following the ideas of tracking by re-detection, proposed in~\cite{bergmann2019tracking}, which emphasizes performing detection and tracking simultaneously. We adapt this idea on the SSD detectors for constant runtime regardless of the number of objects in each frame. 

The tracklet generation stage starts with detecting objects in the first frame and propagating the detection results to the following frames to \textbf{re-detect} these objects. 
Assuming no significant motion exists between two consecutive frames, the method can establish tracklets without additional matching.
Unlike in~\cite{bergmann2019tracking}, where Faster R-CNN ~\cite{ren2015faster} is used and re-detection is implemented by passing the previous frame detection as additional \textbf{tracking proposals} for the next frame, we implement the idea of redetection with single shot face (SSD) detectors. 
This choice is based on the low-latency requirement of video face understanding. 
However, this choice proves to be non-trivial due to an obvious issue. That is, in SSD detectors, there is no proposal classification and regression stage as in two-stage detectors (\emph{e.g.} Faster R-CNN~\cite{ren2015faster}) where the proposals can be passed through.
We solve this problem by proposing a \textbf{tracking anchor assignment} module that emulates the process of adding tracking proposals to two stage detectors. It is inspired by the fact that SSD detectors implement the concept of proposals in a sliding window manner, \emph{i.e.} the anchors. We utilize the fact that a detected bounding box in one frame can still be approximated by a combination of one or multiple anchors, and thus be used for re-detection in the next frame.
For the tracklet linking stage we propose to use a simple online linking algorithm that links a new tracklet to existing tracklets according to its appearance features. 
Thanks to the high-quality tracklets generated by the first module, we found this simple linker can work well in experiments, while at the cost of minimal increase in overall runtime of the SMOT tracker.

We benchmark the proposed \ssfp~ method on three multi object tracking datasets: MusicVideos~\cite{zhang2016tracking}, Hannah~\cite{ozerov2013evaluating}, and MOT17~\cite{milan1603mot16}.
Experimental results show that the \ssfp~can perform fast, accurate, and robust long-term tracking of objects in videos. Especially it significantly reduced the number of transfers~\cite{ristani2016performance}, \emph{i.e.}, the cases of mis-matching multi objects into one track, which is important when there are downstream tasks like recognition relying on the tracking results. 
To summarize, the contributions of this work are:
1) we propose a compact design of multi-object tracking approach that unifies single-shot multi-box (SSD) detector and the tracking-by-redetection scheme;
2) we propose the tracking anchor assign technique to enable propagation of temporal information in SSD detectors;
3) extensive experiments demonstrate that the \ssfp~tracker assisted with a light-weight tracklet linking module can achieve comparable or better performance to the state-of-the-art trackers.

\begin{figure*}[thpb]
      \centering
      \includegraphics[width=0.9\linewidth]{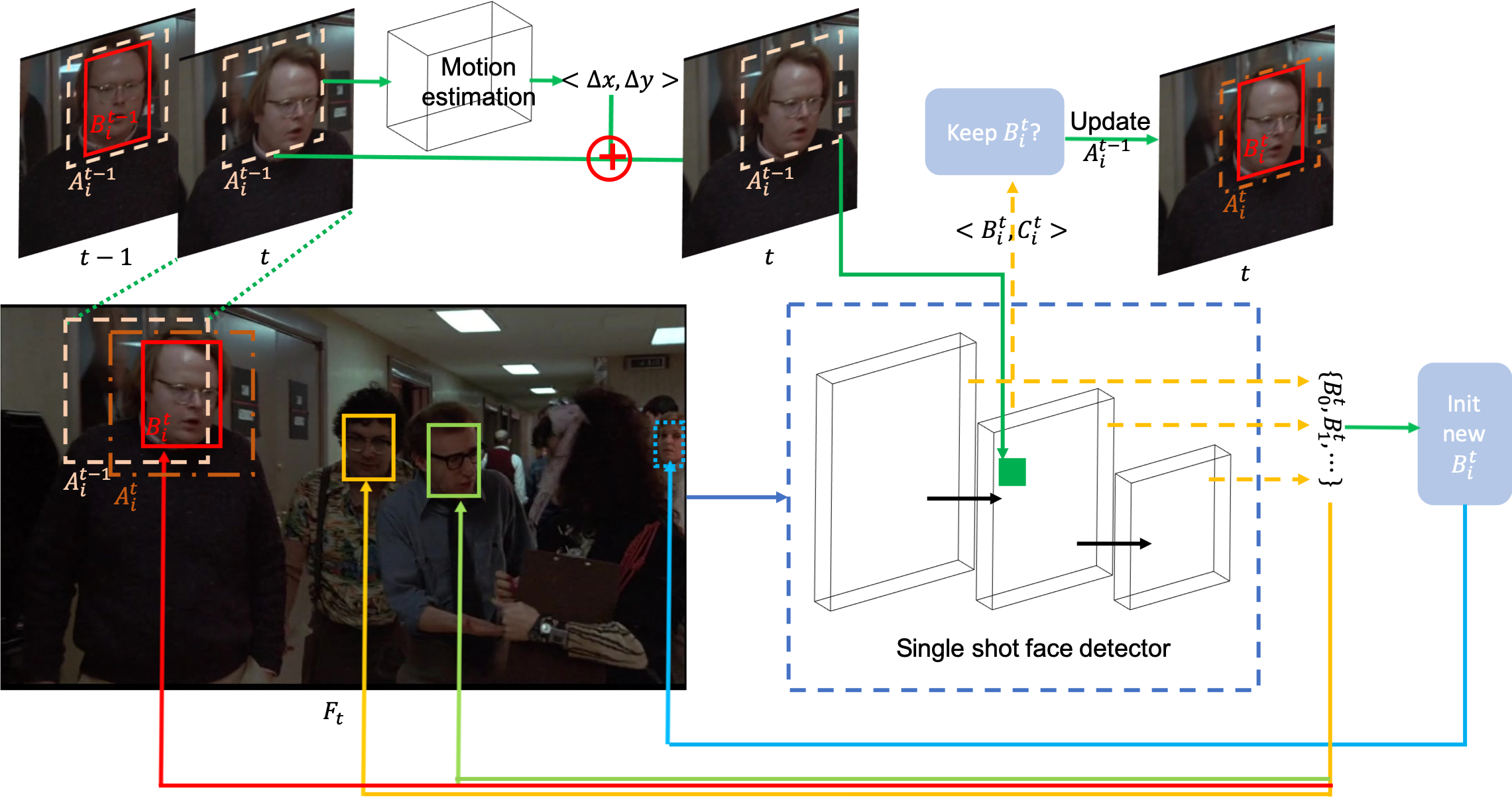}
      \centering
      \caption{Given a frame $F_t$, we first use motion estimator to align existing anchor box $A_i^{t-1}$ associating with the identity $i$ to a new location in the corresponding single shot detector feature map. The regressor and classifier is then used to generate bounding box prediction $B_i^t$ and confidence score $C_i^t$. The $C_i^t$ is used to kill potentially occluded tracks and the $B_i^t$ is used to update the anchor box $A_i^t$. At the same time, the face detector provides a set of detections. We initialize a new track, if a detection has no substantial intersection over union with any bounding box in the set of active tracks. (view in color is suggested)}
      \label{pipeline}
\vspace{-15pt}
\end{figure*}

\section{Related Works}
\textbf{Deep learning-based object detection}
Object detection is a basic computer vision problem. Based on whether using explicit detection proposals, the deep learning detectors can be divided into two main groups: two-stage detectors ~\cite{lin2017feature,ren2015faster,girshick2015fast,he2017mask}  and one stage detectors ~\cite{liu2016ssd,redmon2016you,redmon2018yolov3}. The two-stage detectors first generate detection proposals and then conduct classification as well as location regression. One known issue for the two stage detectors is that the runtime increase dramatically when there is multi
instances to detect, affecting the practical usage in real scenario. While single-shot detectors like SSD ~\cite{liu2016ssd} conduct the proposal generation in an implicit manner. Anchors are generated directly from feature maps and then regressed to the ultimate detection results. The anchor-based regression is agnostic to instance numbers so it could be well applied in scenes where multi-target detection is required like face detection ~\cite{deng2019retinaface}. 

\textbf{Multi-object tracking}
Benefiting from the advance of object detection, we saw the emergence of many multi-object tracking approaches ~\cite{voigtlaender2019mots,xu2019deepmot,bochinski2017high,insafutdinov2017arttrack,ma2018trajectory}. ~\cite{wang2019exploit} proposed a multi-scale TrackletNet to merge tracklets generated through CNN features, IoU scores and  epipolar geometry.  ~\cite{feichtenhofer2017detect} combined the tracking and detection, providing a simultaneous approach as detection shows the power in regressing accurate position of the targets. The author developed the D2T framework by adding a correlation layer to a Siamese R-FCN \cite{dai2016r} detector. Global tracklets fusion are employed later to generate final tracking results. Compared to these batch based tracking methods, online multi-target tracking has drawn more and more attention due to its run-time potential in practice. ~\cite{Wojke2018deep} provides a simple but effective online multi-target solution. To realize online multi-target tracking, the author combined the deep learning based detection, feature embedding with traditional MOT framework and proposed deep feature \& IoU merging based two-gate tracking methods. Different from these tracking by detection methods where complete detection are employed at the first stage, ~\cite{bergmann2019tracking} builds the connection between tracking and detection in a more entangled way. It saves detection results and transforms as new detector proposal to regress target positions in the new frame. This approach has a strong assumption that the target does not have large motion, and thus the new regressed position and the proposal corresponds to the same object. This means the pair of bounding boxes automatically form a tracklet without any additional association efforts.

\section{Pipeline Overview}
Before diving into details of each stage in the \ssfp~, we first review the overall object tracking pipeline.
The input to the \ssfp~is a video, represented by a consecutive sequence of $T$ frames, $\{F_i\}_{t=1}^T$, and the output is a series of tracklets $\{S_i\}$ representing object locations from the input sequence.
Here one tracklets is represented by a set of object bounding boxes $B_i^t$ from frames with the same pathing ID $i$. 
And each bounding box is in the format of: top-left corner $(x_i^t, y_i^t)$, height $h_i^t$, and width $w_i^t$.

\ssfp~operates with a cascade of two components, as illustrated in Fig.~\ref{fig:teaser}.
The first stage is called \emph{tracklet generator}, in the sense that it focuses on generating tracklets from the video frames.
A tracklet $L_j$ is a short-term consecutive bounding box sequence, which encapsulates a series of bounding boxes of the same object instance in several consecutive frames. 
In \ssfp~ the tracketlet generator is converted from a normal SSD detector. 
At timestep $t$, the tracklet generator observes the current frame $F_t$.
It also carries the tracklets that were present in the previous frame. 
We detect the known objects again in the new frame by propagating their information from the previous frames to extend the tracklets. We call this operation \emph{re-detection}.
Re-detected objects are used to extend the existing tracklets by inheriting their identity. 
Existing tracklets failed to be re-detected will be terminated and released from the tracklet generator. 
The tracket generator also performs normal detection on the new frame. 
The result bounding boxes will first be merged with existing tracklets to avoid duplication, where the remaining standalone bounding boxes will be then used to initiate new tracklets.
The second stage is called \emph{tracklet linking}, which works in an online manner to link tracklets into long-term ones. 
The tracklet linking extracts appearance embedding feature for a newly formed tracklet and matches it with the existing tracklets. 
The matched tracklets are then linked to the corresponding tracks with the same track ID. 
Non-match tracklets will start as new tracks with their own new track ID.
In the following sections, we will go through the details of the two stages.

\section{Tracklet Generator}\label{sec:tracklet_generator}
The trackelet generator is based on the idea of tracking by redetection, which performs detection and tracking simultaneously.
This scheme is recently proposed by~\cite{bergmann2019tracking} for the task of person bounding box pathing. 
In this work, the basic motivation is that the bounding box regressors in two-stage object detectors, such as faster R-CNN, has the capacity of rectifying the locations and shapes of rough object proposals. 
Thus by propagating the rough locations of the tracked objects in the previous frames, we can re-detect these objects and naturally extend the tracklets overtime.
In~\cite{bergmann2019tracking}, this idea is implemented on a faster R-CNN~\cite{ren2015faster} person detector trained on the COCO dataset~\cite{lin2014microsoft}.
The bounding boxes of the tracked objects in the previous frame are propagated to the next frame as additional ``tracking proposals'' for classification and regression. 
It is proven to be effective for the person tracking task. 
However, there is a downside of this implementation, that its runtime will grow linearly with the number of objects, which becomes a problem for multi-object tracking in crowded scenes where a fast, constant runtime is required.
In \ssfp, we design the tracklet generator following the tracking by redetection scheme, but using a single shot face detector ~\cite{liu2016ssd} (SSD) as the backbone. 
The goal is to have an almost constant runtime regardless of the number of objects in the frames. However, it is non-trivial to integrate the tracking by redection scheme to SSD detectors, which have no proposal classification and regression process as in Faster-RCNNs~\cite{ren2015faster}, where we can insert the tracking proposals . 
To solve this issue, we propose a novel \textbf{tracking anchor assignment} module, which enables the process of propagating tracked object locations between frames without explicit proposals.

\subsection{Redetection with Tracking Anchors}\label{sec:redect}
Unlike Faster-RCNN, SSD detectors do not have an explicit concept of proposals. Instead, they operate by classifying and rectifying the locations of a set of predefined location hypothesis, \emph{i.e.}, anchors.
Therefore, another way to view single shot detectors is that they actually have a fixed number of proposals in the form of anchors with prior locations and shapes.
The network then, based on the image features, makes prediction for each proposal. 
This gives us a way to reinstate in the current frame the bounding boxes from tracked object in previous frames. 
Instead of inserting tracked object locations as proposals, we can use a set of predefined anchors to approximate the rough locations of the tracked objects. 
We call this set of anchors the \textbf{tracking anchors} of an object. 
By aggregating the detection output of the tracking anchors, we can then estimate the redetection result as if the tracked object locations were inserted as proposals.

Formally, assuming for a frame $F_{t}$, we have detected several object bounding boxes assigned to tracklet $\{L_j^t\}$.
For frame $F_{t+1}$, we adopt the following steps to re-detect the objects.
For each bounding box $B_j^t$ of the trackle $\{L_j^t\}$, we predict its location and shape at $t+1$ by function $\hat{B}_j^{t+1} = H(B_j^t)$.
The simplest form of $H$ can be the identity function, which means 
$\hat{B}_j^{t+1} = B_j^t$.
From all the detector anchor candidates, we extract a set of $K$ anchors $\{a_{jk}\}_{k=1}^{K}$ which are close to $B_j^t$ in both scale and location spaces. 
This set of anchors will serve as the tracking anchors of $B_j^t$. They each carries a weight coefficient $w_{jk} = A(a_{jk},\hat{B}_j^t) $ determined by the tracking anchor assignment function $A$. 
The redetection can then be performed by aggregating the detection output of these anchors the bounding boxes shapes $\{B_{jk}^{t+1}\}$ and the detection confidence scores $\{c_{jk}^{t+1}\}$ on frame $F_{t+1}$, defined as following: 
\begin{align}\label{eq:redetect}
  B_j^{t+1} = \sum_{k=1}^{K}\frac{w_{jk}^{t+1} B_{jk}^{t+1}}{w_{jk}^{t+1} },\quad
  c_j^{t+1} = \sum_{k=1}^{K}\frac{w_{jk}^{t+1} c_{jk}^{t+1}}{w_{jk}^{t+1} },
\end{align}
where $B_j^{t+1}$ is the redetected bounding box shape of the tracked object $j$ in frame $F_{t+1}$ and $c_j^{t+1}$ is its confidence score.
Determined by a tracking specific threshold $\sigma_{active}$, there are two possible outcomes for these anchors:
a) if $c_j^{t+1} \geq \sigma_{active}$, this is a valid redetection. The bound box $B_j^{t+1}$ is considered as the new presence of the tracked object $j$;
b) if $c_j^{t+1} < \sigma_{active}$, this is a failed redetection. The tracklet is considered terminated, and released from the tracklet generator.

\noindent\textbf{Tracking anchor assignment strategy}.
In the \ssfp~framework, we use strategies to build the tracking anchor assignment module. The simplest strategy is when $K=1$, we just take the anchor ${a_j^{t+1}}$ that has the highest intersection over union (IoU) with $\hat{B}_j^{t+1}$. In experiments, we found that this simple strategy, known as ``single-assignment'', performs well for the task like face or person tracking.
When $K > 1$, we can take the top-$K$ highest IoU anchors with $\hat{B}_j^{t+1}$ and use their IoU values as the weights. We call this strategy ``multi-assignment''.
It is also possible to use other strategies such as learned weighting, which we leave as future work.  

\begin{figure}[thpb]
      \centering
      \includegraphics[width=1\linewidth]{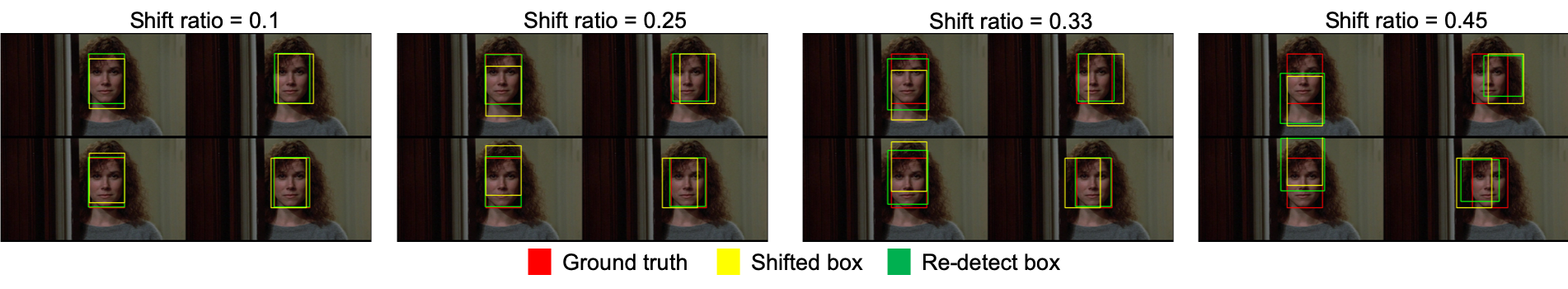}
      \centering
      \caption{Robustness test of the anchor regression (a-d corresponds to shift ratio of 0.1, 0.25, 0.33, 0.45, respectively). Red denotes the ground truth, yellow denotes the shifted box and green the re-detected target position (better view in color ).}
      \label{fig_aug_ratio}
\end{figure}

\noindent\textbf{Is the prediction by the tracking anchors reliable?}
To answer this question, we conduct a toy experiment on a small set of images. 
We look at the predictions of the anchors that are similar in terms of intersection over union (IoU) to the ground truth bounding boxes. 
We found most of these anchors' predicted bounding boxes point to the groundtruth bounding boxes.
Quantitatively, we demonstrated in Fig.~\ref{fig_aug_ratio} how the initial shift of a bounding box affect the ability to correctly detect an object.
Using the $w,h$ of the bounding box in $F_{i}$ as a scale, we added multi ratios of shift to the image for re-detection. Fig.~\ref{fig_aug_ratio} shows the re-detection results. We found that the detector can correctly localize the object position even with a relative shift ratio of 0.33. This means if there is only a small motion between two consecutive frames, we can accurately track the objects in the new frames.
Similar to their counterpart in faster RCNN, the bounding box regressors in SSD also have strong power to rectify anchors, producing tight bounding boxes of the objects.
If the tracking anchors are close to the rough locations of the object to track, there is a high chance to correctly find its new location, so we can rely on the aggregation of these tracking anchors' predictions for redetection.

\noindent\textbf{The workflow of tracklet generation}. In summary, we can describe the workflow of our tracklet generator as following:

(1) For the first frame $F_0$: run the single shot object detector and get all object detection results $\{B_i^0\}$. Create one tracklet ID for each bounding box;

(2) Predict the locations and shapes of these tracked objects $\{\hat{B}_j^{1}\}$ in frame $F_{1}$ . Obtain the tracking anchors $\{a_{ik}^0\}$ and their weights $\{w_{ik}^0\}$ for each object $B_j^1$ through the tracking anchor assignment module;

(3) Run the detector on frame $F_1$. Obtain the redetection results according to Eq.~\ref{eq:redetect}, decide whether to extend or halt the tracklet according to the tracking specific threshold $\sigma_{active}$;

(4) Perform non-maximal suppression on the re-detection bounding boxes to merge overlapping tracklets, by suppressing new detection results that have high IoU to the tracked bounding boxes.

(6) For remaining new detection results, create a new tracklet for each of them.

(7) Repeat the steps 2-6 for the remaining frames consecutively.

\subsection{Motion Assisted Re-Detection}
Another important module in the \ssfp~tracklet generator is the function $H$ that predicts the rough location of a tracked object in the new frame. The simplest form of $H$ is the identity function, which implies a small motion assumption. 
When the video contains large motion, this assumption breaks and it becomes easy for the tracklet generator to lose the target. 
Therefore, it is important to consider motion in bounding box prediction. 

Motion estimation from two consecutive frames is a widely investigated problem~\cite{ilg2017flownet,ilg2018occlusions}. 
Dense optical flow algorithms provide pixel-wise displacement prediction between two frames, which can be used to predict the bounding box movement. 
We propose to use the dense optical flow field to build the function $H$ for each tracked object.
Particularly, given the optical flow field $D_t$ between $F_{t}$ and $F_{t+1}$ and a two element displacement vector (horizontal and vertical) $\Vec{d}_{(h, w)}$ on each pixel location, we can predict the new location of the bounding box $B_j^{t}$ in frame $F_{t+1}$ as
\begin{align}
    \hat{B}_j^{t+1} = H_{motion}(B_j^{t}) = B_j^{t} \oplus \frac{\sum_{(h, w) \in B_j^{t}} \Vec{d}_{(h, w)}}{\mathrm{area}(B_j^{t})},
\end{align}
where the operator $\oplus(B, \Vec{d})$ shifts the center of the bounding box $B$ by the vector $\Vec{d}$ and the function $\mathrm{area}(B)$ gives the area of the bounding boxes in square-pixels.
We call the tracklet generation process aided with this form of $H$ as motion-assisted redetection.
In experiments, we compare two different dense flow estimation approaches, a traditional FarneBack flow~\cite{farneback2003two} and another deep learning based PWCNet flow~\cite{sun2018pwc}.
We found that the motion assisted redetection can benefit from both approaches even though their optical flow estimation benchmark numbers are quite different.

\begin{figure}[thpb]
      \centering
      \includegraphics[width=1\linewidth]{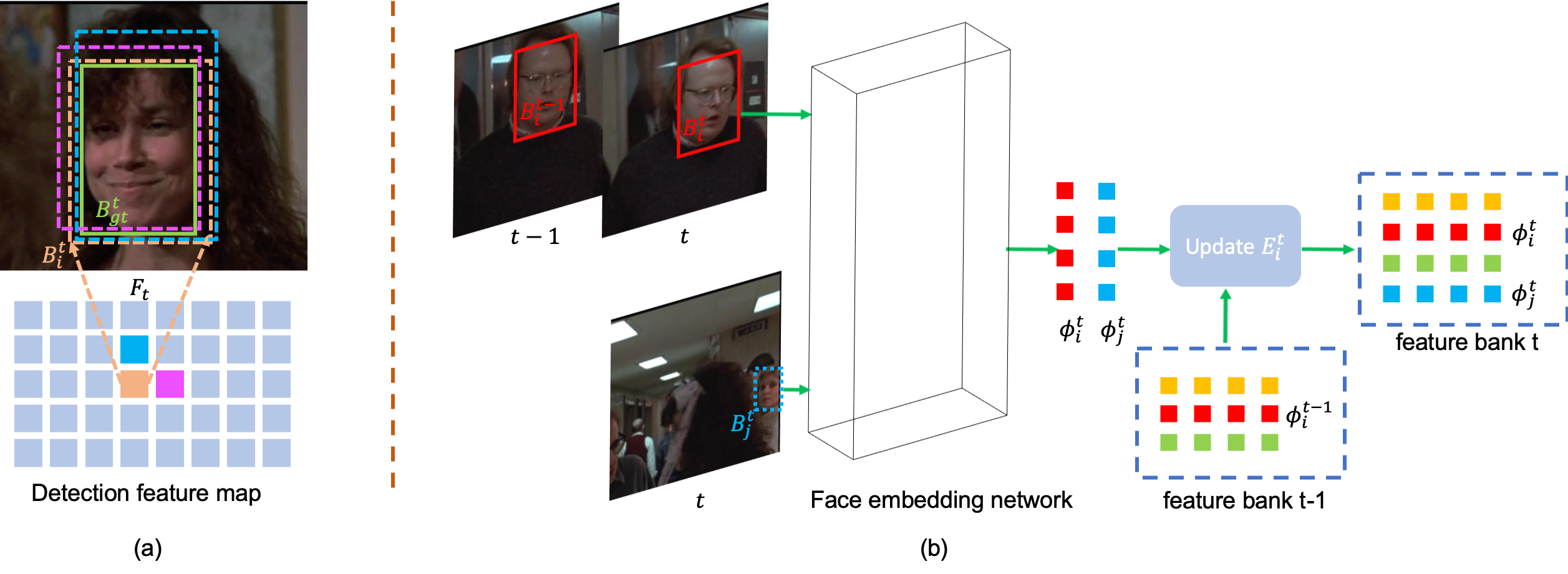}
      \centering
      \caption{Online updating of the feature bank for tracklet-track linking. $\phi_i^t$ denotes the identity embedding of track $i$ at time $t$. We store a bank of the identity embeddings where each track has one feature. When we have a new tracklet, we either match it to an existing identity in the bank or add a new one to the bank (better view in color).}
      \label{linking}
\end{figure}
\vspace{-15pt}
\section{Online Linking of the Tracklets}
Since the output of the tracklet generator are short tracklets, which are usually halted due to occlusion and temporary disappearance of objects in videos,
we design an online tracklet linking module to group the tracklets into final object tracks.

The linker carries a set of existing tracks $\{S_i\}$, where each track is represented by an identity embedding $\phi(S_i)$ produced by averaging the output of a embedding model $\phi$ on objects in this track.
At a certain time step $t$, there are two possible states of a track: 1) this track is linked to an active tracklet in the current frame $F_t$; 2) this track has no active tracklet. Fig. ~\ref{linking} illustrates the identity linking process.
For each new tracklet $L_j^t$ appearing in $F_t$, we try to match it with tracks in state 2). 
To perform the matching, we first extract the appearance embedding from the object in $L_j^t$, then we calculate the pairwise match cost between $L_j^t$ and the tracks. Once a track has a new tracklet linked to it, it goes to state 1) in the next frame. By solving the bipartite matching problem we can link the matching tracklet to the tracks. For those new tracklets that are not matched to any existing tracks, generate a new track for each of them.

This tracklet linker handles the cases of occlusion and temporal disappearance across time. 
On the other hand it requires extracting appearance embedding using an embedding model, usually a deep neural network like~\cite{deng2019arcface,shen2017learning}, which adds to the runtime cost.
However, this additional cost can be minimal due to two reasons:
1) the linking is only performed for newly created tracklets, once a tracklet is linked we do not need to re-extract its embedding.
2) the extraction of appearance embedding for the track do not need to be run on each frame, as in videos the appearance of objects usually change slowly.

\section {Experiments}
We evaluate the proposed \ssfp~tracker on three multi object tracking datasets: the Hannah dataset~\cite{ozerov2013evaluating}, the Music Videos dataset~\cite{zhang2016tracking}, and the MOT17~\cite{milan1603mot16} dataset.
The approach is assessed in term of standard multi object tracking metrics, runtime speed, and metrics related to downstream tasks such as video face recognition or person identification.
We also provide qualitative analysis of the approach's ability in handling common nuance factors to object tracking.
\subsection{Implementation details}
In our implementation of \ssfp, we implement the single shot object detector with ResNet34~\cite{he2016deep} 
~as the backbone architecture for both person and face tracking tasks. Feature pyramid network~\cite{lin2017feature} is adopted to enhance the context modeling ability.
We set the confidence threshold $\sigma_{det}$ to $0.9$ for detection and $\sigma_{tk}$ to $0.4$.for tracking.
We set the NMS threshold for redetection as $0.6$ and the IoU threhold between tracks and new detection as $0.3$.
In motion assisted re-detection, we set optical flow field size for the Farneback algorithm to $256\times 256$ pixels and $1024\times1024$ for the PWCNet~\cite{sun2018pwc}, respectively.
We use the face embedding network with 128-dimensional feature vectors proposed in~\cite{deng2019arcface} for tracklet linking module and the threshold distance of face verification is 0.97. No appearance model and tracklet linker is used for person tracking. The tracklet generator in the person tracking task uses $0.9$ as detection threshold and $0.4$ as tracking threshold to keep high precision of our bounding box quality, as well as better recall capability for tracking regression.
\subsection{Dataset}
 To evaluate the performance of our approach as generic multi-object tracking, we test on two common use cases: person tracking and face tracking. We test our approach in the MOT17 dataset to see how it performs in person tracking, while on Hannah and MusicVideo dataset for face tracking scenario.

\noindent\textbf{MOT17} ~\cite{milan1603mot16} is a challenging dataset designed for multi object tracking. The dataset includes 7 video sequences for training and 7 videos for testing, covering indoor and outdoor person tracking cases. There are lots of person occlusion, fast motion, making the tracking even more challenging. 

\noindent\textbf{Hannah dataset}~\cite{ozerov2013evaluating} is a public available face tracking dataset from the Hollywood movie \textit{Hannah and her sisters}. There are in total 153825 frames and 245 shots. As a multi face tracking dataset, it contains 254 people and 53 of whom are named characters. Overall 2002 tracks are annotated, ranging from 1 to 500 frames in length. 

\noindent\textbf{Music video dataset} ~\cite{zhang2016tracking} is a widely used face tracking benchmark with $8$ high resolution music videos from YouTube ranging from four to six minutes in duration. The videos contain
edited scenes with singing and dancing movement. Each video contains around 3000 to 5000 frames with 1 to 10 people in the video. It is a challenging benchmark for face tracking due to fast motion, shot changing and crowd occlusions.

\vspace{-15pt}
\subsection{Evaluation Metric}
For the multi object trackings tasks we report the performance metrics as discussed in~\cite{ristani2016performance}. 
In particular, we report multi-object tracking accuracy (\textbf{MOTA}), min-cost matched f1 scores (\textbf{IDF1})), recall, precision, f1 score (\textbf{F1}), number of transfers (\textbf{Transfer}), and id switches (\textbf{IDsw.}).

\subsection{Baseline approaches} We implemented the widely used DeepSORT tracker~\cite{Wojke2018deep} as one of our baseline approaches. It is based on the standard tracking by detection scheme. 
As shown in ~\cite{wojke2017simple}, the performance of the DeepSORT tracker is highly dependent on the quality of its object detector and re-ID features.
For face tracking, we equip this approach with the state-of-the-art image based face detector trained on the WIDER Face~\cite{yang2016wider} dataset and the state-of-the-art face embedding~\cite{deng2019arcface} for re-identification features, same to our approach for fair comparison. We refer to this method as \textbf{DFS} (Deep Face Sort).
Another baseline method we employed is the Tacktor++ described~\cite{bergmann2019tracking}. 
For face tracking benchmarks, we use a face detector based on faster-RCNN+FPN and trained on the WIDERFace~\cite{yang2016wider} dataset for fair comparison.
For person tracking, our model detector is trained on COCO ~\cite{lin2014microsoft} and finetuned on MOT17 ~\cite{milan2016mot16} training split. And then benchmark the result on MOT17 under the same experiment protocol with Tracktor++~\cite{bergmann2019tracking}.

\begin{figure}[thpb]
      \centering
      \includegraphics[width=.9\linewidth]{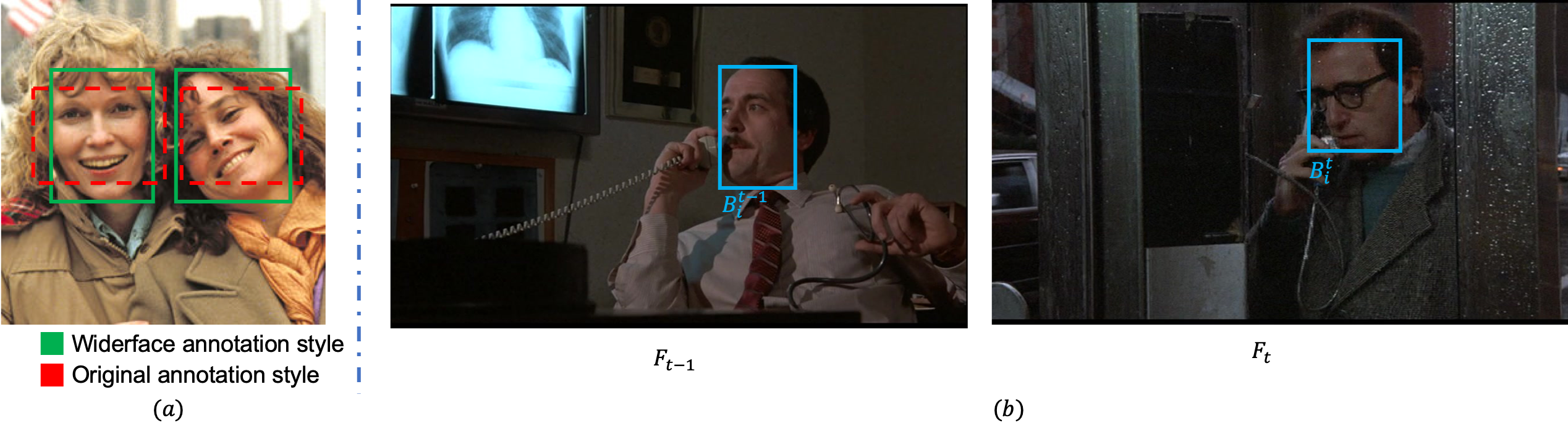}
      \centering
      \caption{Fig. $(a)$ shows the different bounding box styles of Hannah dataset and WiderFace, WiderFace has tight bounding box covering the face region while Hannah style bounding box is wider horizontally including ears but shorter in vertical which starts from forehead center to jaw center. Fig. $(b)$ illustrates the rare case where shot changes but new face appear at the same region, this leads to identity miss match in tracking}
      \label{fig_dataset}
\end{figure}

\vspace{-15pt}
\subsection{Comparison with State of the Arts}
Table ~\ref{tab:tab_overall} shows the evaluation results on the three benchmarks. We compare our basic tracklet generator, motion-assisted tracklet generator and the full \ssfp~tracker with the state-of-the-art approaches. We can observe that our basic tracklet generator shows much better performance in number of transfers and recall rate compared to the DFS methods, which means it miss-matches less different objects and detects more object boxes with the help of temporal propagation.
We also compared with the tracktor++ baseline results, we find that our tracklet generator shows comparable or better tracking accuracy than tracktor++. Furthermore for the setting of online tracking, the \ssfp outperforms the the state-of-the-art-methods. Even compared with other offline approaches where more spatial and temporal information are fused to improve tracking \cite{lin2018prior} \cite{zhang2016tracking}, the method has comparable performance and shows significant improvement in metric like transfer.   
This demonstrates the effectiveness of our solution, employing tracking anchor assignment and aggregation in single-stage tracking.

\vspace{-15pt}
\subsection{Ablation Study}

\noindent\textbf{Detection and tracking thresholds.}
We first examine the impact of the two confidence thresholds in the tracklet generator, namely, the detection threshold and the tracking threshold $\sigma_{active}$. 
The results on both datasets are shown in Table~\ref{tab_anchor} where $d$ and $t$ stands for detection and tracking confidence threshold, respectively.
We can observe that generally a higher detection threshold leads to fewer false positive detections and thus improves the precision rate, while 
the recall rates are less affected, so we can choose a high detection threshold for the tracklet generator. 
The number of transfers can be further reduced by tuning the tracking threshold $\sigma_{active}$ .
We recommend the combination of $0.9$ and $0.4$ for the detection and tracking thresholds.

\noindent\textbf{Tracking Anchor Assignment Strategies}
In sec~\ref{sec:redect}, we introduced the tracking anchor assignment module. It is flexible with multi assignment strategies. Here we experimented with two of them. The first is the single-assignment where only one anchor $i$ is assigned for each tracked object. The second is the multi-assignment strategy, where a set of $K$ anchors is used and their IoU scores with the tracked objects are used as aggregation weights. 
The results are shown in Table~\ref{tab_anchor}. 
We observe that even with one single tracking anchor, the tracklet generator can still work quite well. 
With $K$ increasing, the tracklet generator is able to generate longer tracklets with fewer breaks, which is illustrated by the increasing IDF1 and the decreasing ID switches.

\noindent\textbf{Dense optical flow algorithms.}
The motion assisted redetection involves computation of dense optical flow. 
We experimented with two types of algorithms , the first is FarneBack~\cite{farneback2003two} algorithm, and the second is the PWCNet~\cite{sun2018pwc}. Here the baseline is the basic tracklet generator without motion assistance.
We found that involving motion can significantly improve the performance. And the more accurate PWCNet leads to better tracking accuracy. This suggests that with better learned motion prediction module, the tracklet generator could be further improved.

\begin{table}
\centering
\tiny
\caption{\ssfp~ tracker performance evaluation}
\label{tab:tab_overall}
\begin{tabular}{ |c|c|c|c|c|c|c|c|c|c|c| }

\hline
&\multicolumn{10}{c|}{MOT17} \cr
\hline
      Methods & MOTA & IDF1 &ID Precision &ID Recall & Transfer & F1& Precision &Recall & ID switch & Online?\\
\hline

  Tracktor++ ~\cite{bergmann2019tracking} &0.588 &0.623 &- &- &- &0.761&0.99 &0.626 &1425  & \checkmark \\
  \ssfp &0.529 &0.477 &0.639 &0.382 &690 &0.721 &0.965 &0.576 &2491 & \checkmark\\
  \ssfp w/ Farneback& 0.554 &0.538 &0.731 &0.426 &89 &0.721 &0.979 &0.571 &743 &\checkmark \\
  \ssfp w/ PWCNet& 0.56 &0.539 &0.721 &0.43 &69 &0.728 &0.974 &0.581 &620  &\checkmark \\

\hline

\hline
&\multicolumn{10}{c|}{Hannah} \cr
\hline
      Methods & MOTA & IDF1 &ID Precision &ID Recall & Transfer & F1& Precision &Recall & ID switch & Online?\\
\hline
  DFS ~\cite{Wojke2018deep}& 0.642 &0.671 & 0.604 &0.753 &71 &0.842 &0.759 &0.946 &788 & \checkmark \\ 
  Tracktor++ ~\cite{bergmann2019tracking} &0.542 &0.552 &0.484 &0.642 &46 &0.807 &0.708 &0.938&2042  & \checkmark \\
  \ssfp & 0.609 &0.569 &0.507 &0.646 &20 &0.784 &0.698 &0.896 &3449 & \checkmark\\
  \ssfp w/ Farneback& 0.726 &0.580 &0.569 &0.590 &10 &0.870 &0.853 &0.885 &1499 &\checkmark \\
  \ssfp w/ PWCNet& 0.716 &0.565 &0.555 &0.574 &8 &0.865 &0.851 &0.880 &1805  &\checkmark \\
  \ssfp w/ online linking& 0.728 &0.654 &0.642 &0.667 &165 &0.869 &0.853 &0.885 &1122 &\checkmark \\
\hline
&\multicolumn{10}{c|}{MusicVideo} \cr
\hline
  Siamese~\cite{zhang2016tracking} & 0.623 &- &- &- &986 &0.795 &0.894 &0.715 &- &  x \\ 
  SymTriplets~\cite{zhang2016tracking}& 0.638 &- &- &- &699 &0.798 &0.897 &0.718 &- & x \\ 
  Prior-Less~\cite{lin2018prior} &0.692 &- &- &- &624 &0.853 &0.902 &0.817 &- & x \\ 
  DFS~\cite{Wojke2018deep}&0.614 &0.141 &0.141 &0.142 &400 &0.84 &0.842 &0.837 &5641 &\checkmark \\ 
  Tracktor++~\cite{bergmann2019tracking}&0.417 &0.117 &0.109 &0.127 &230 &0.765 &0.709 &0.830 &6537 & \checkmark \\
  \ssfp &0.536 &0.114 &0.106 &0.123 &38 &0.831 &0.772 &0.900 &8866 & \checkmark\\
  \ssfp w/ Farneback&0.571 &0.120 &0.118 &0.121 &44 &0.821 &0.809 &0.834 &5693 &\checkmark \\
  \ssfp w/ PWCNet&0.552 &0.111 &0.099 &0.127 &49 &0.803 &0.813 &0.824 &5894 &\checkmark \\
  \ssfp w/ online linking&0.615 &0.244 &0.232 &0.259 &545 &0.853 &0.808 &0.904 &6525 &\checkmark \\
\hline

\end{tabular}
\vspace{-15pt}
\end{table}

\begin{table}
\centering
\tiny
\caption{\ssfp~ tracker performance with different detection \& tracking confidence thresholds.}
\label{tab_anchor}
\begin{tabular}{ |c|c|c|c|c|c|c|c|c|c|c|c|c| }

\hline
&\multicolumn{6}{c|}{Hannah} &\multicolumn{6}{c|}{MusicVideo} \cr
\hline
      $\sigma_{det}$ $\sigma_{tk}$  & MOTA & IDF1 & Transfer & F1&Precision & ID switch& MOTA & IDF1 & Transfer & F1&Precision & ID switch \\
\hline
0.5 0.1 & 0.468 & 0.542  & 22 & 0.786 & 0.671  & 3122 & 0.416 & 0.108 & 51  & 0.698 & 0.784   & 8400 \cr  
0.65 0.3 & 0.552 & 0.541  & 13 & 0.819& 0.716  & 4994 & 0.493 & 0.109 & 32  & 0.751 & 0.824   & 10673 \cr  
0.8 0.2 & 0.573 & 0.559  & 20 & 0.823 & 0.725  & 3558 & 0.511 & 0.112 & 38  & 0.756 & 0.823   & 9066 \cr 
0.8 0.3 & 0.565 & 0.513  & 15 & 0.837 & 0.746  & 12106 & 0.465 & 0.110 & 32  & 0.773 & 0.834   & 15629 \cr 
0.8 0.4 & 0.590 & 0.532  & 12 & 0.837 & 0.745  & 7330 & 0.512 & 0.111 & 25  & 0.779 & 0.841   & 12722 \cr 
0.9 0.2 & 0.609 & 0.569  & 20 & 0.836 & 0.746  & 3449 & 0.536 & 0.114 & 38  & 0.772 & 0.831   & 8866 \cr 
0.9 0.3 & 0.607 & 0.524  & 15 & 0.851 & 0.771  & 11579 & 0.493 & 0.112 & 32  & 0.791 & 0.841   & 14988 \cr 
0.9 0.4 & 0.627 & 0.543  & 11 & 0.849 & 0.768  & 6885 & 0.538 & 0.113 & 25  & 0.796 & 0.848   & 3122 \cr  \hline
\end{tabular}

\vspace{-5pt}
\end{table}

\begin{table}
\centering
\tiny
\caption{\ssfp~ tracker performance v.s. numbers of anchor aggregation}
\begin{tabular}{ |c|c|c|c|c|c|c|c|c|c|c| }

\hline
&\multicolumn{5}{c|}{Hannah} &\multicolumn{5}{c|}{MusicVideo} \cr
\hline
      Num of anchor & MOTA & IDF1 & Transfer & F1 & ID switch&MOTA & IDF1 & Transfer & F1 & ID switch \\
\hline
1 & 0.612 & 0.573  & 13 & 0.837  & 3239 &  0.563  & 0.116 & 27 & 0.845 & 8821  \\ 
5 & 0.607 & 0.581  & 17 & 0.835  & 3040 &  0.553  & 0.117 & 34 &0.837 & 8361 \\
10 & 0.609 &0.569   & 20 & 0.836  & 3449 & 0.536 & 0.114 & 38 & 0.831  & 8884 \\
20 & 0.608 & 0.581  & 17 & 0.835  & 3040 &  0.553  & 0.117& 34 & 0.837  & 7495 \\
\hline
\end{tabular}
\vspace{-5pt}
\end{table}

\begin{table}[h!]
\centering
\tiny
\caption{\ssfp~tracker runtime v.s. performance}
\begin{tabular}{ |c|c|c|c|c|c|c|c|c|c|c|c| }

\hline
& &\multicolumn{5}{c|}{Hannah} &\multicolumn{5}{c|}{MusicVideo} \\
\hline
      Model &FPS& MOTA & IDF1 & Transfer & F1 & ID switch&MOTA & IDF1 & Transfer & F1 & ID switch \\
\hline
Tracktor ++ 1024& 9 & 0.542 & 0.552  & 46 & 0.807  & 2042 &  230  & 0.117 & 230 & 0.765 & 6537\\
\ssfp 1024&17 & 0.609 & 0.569  & 20 & 0.836  & 3449 &  0.537  & 0.114 & 38 & 0.832 & 8866  \\ 
\ssfp 768&22 & 0.612 & 0.568  & 21 & 0.837  & 3443 &  0.536  & 0.114 & 39 &0.829 & 8746 \\
\ssfp 512&27 & 0.621 &0.566   & 20 & 0.840  & 3464 & 0.529 & 0.114 & 36 & 0.826  & 8884 \\
\ssfp 256&40 & 0.708 & 0.536  & 11 & 0.878  & 3477 &  0.506  & 0.114& 35 & 0.798  & 7495 \\
\hline
\end{tabular}
\vspace{-15pt}
\label{runtime_exp}
\end{table}

\vspace{-15pt}
\subsection{Runtime Analysis}
One of the main advantages of \ssfp is its fast speed. We conducted a comparison experiment to explore the runtime efficiency. In the experiment setting, we resize the input image to $1024\times 1024$, $768\times768$, $512\times512$, $256\times256$ respectively. With one NVIDIA V100 GPU on the different resolution settings, the \ssfp can achieve 17, 22, 27, 40 frame per second as shown in Table ~\ref{runtime_exp}, respectively. We also don't find significant change in runt-time when the resolution changes or the number of tracked object increases. This shows the \ssfp is fast in runtime and robust in image resolution changes, which makes it possible for real-time processing in different application scenarios. Note that tracklet generation module is independent to the backbone detector structures. It can be applied to any single shot detector. All the current components can be replaced with faster versions in order to achieve real-time performance in much less powerful machines, like edge devices.

\vspace{-15pt}
\section{Conclusion}
In this paper, we present a multi object tracking approach, named Single-shot Multi Object Tracker (SMOT),  that is able to simultaneously generate detection and tracking outputs. It is composed of a tracklet generator and a tracklet linker. By applying the tracking by re-detection idea to single shot object detectors, the tracklet generator runs at close to constant speed w.r.t. the number of targets. Motion assisted redetection makes it robust to large motion in video tracking cases. A simple but effective tracklet linker merges the tracklets in an online manner and further improves the tracking result.
Extensive experiments shows the SMOT method achieves state-of-the-art accuracy when comparing with other online object tracking approach, with comparable performance even to the state-of-the-art offline methods. 

{
\bibliographystyle{splncs04}
\bibliography{arxiv_main}

\begin{thebibliography}{10}
\providecommand{\url}[1]{\texttt{#1}}
\providecommand{\urlprefix}{URL }
\providecommand{\doi}[1]{https://doi.org/#1}

\bibitem{bergmann2019tracking}
Bergmann, P., Meinhardt, T., Leal-Taixe, L.: Tracking without bells and
  whistles. In: ICCV (2019)

\bibitem{bochinski2017high}
Bochinski, E., Eiselein, V., Sikora, T.: High-speed tracking-by-detection
  without using image information. In: 2017 14th IEEE International Conference
  on Advanced Video and Signal Based Surveillance (AVSS). pp.~1--6. IEEE (2017)

\bibitem{braso2020learning}
Bras{\'o}, G., Leal-Taix{\'e}, L.: Learning a neural solver for multiple object
  tracking. In: Proceedings of the IEEE/CVF Conference on Computer Vision and
  Pattern Recognition. pp. 6247--6257 (2020)

\bibitem{dai2016r}
Dai, J., Li, Y., He, K., Sun, J.: R-fcn: Object detection via region-based
  fully convolutional networks. In: NIPS. pp. 379--387 (2016)

\bibitem{deng2019arcface}
Deng, J., Guo, J., Xue, N., Zafeiriou, S.: Arcface: Additive angular margin
  loss for deep face recognition. In: CVPR. pp. 4690--4699 (2019)

\bibitem{deng2019retinaface}
Deng, J., Guo, J., Zhou, Y., Yu, J., Kotsia, I., Zafeiriou, S.: Retinaface:
  Single-stage dense face localisation in the wild. arXiv preprint
  arXiv:1905.00641  (2019)

\bibitem{farneback2003two}
Farneb{\"a}ck, G.: Two-frame motion estimation based on polynomial expansion.
  In: Scandinavian conference on Image analysis. pp. 363--370. Springer (2003)

\bibitem{feichtenhofer2017detect}
Feichtenhofer, C., Pinz, A., Zisserman, A.: Detect to track and track to
  detect. In: Proceedings of the IEEE International Conference on Computer
  Vision. pp. 3038--3046 (2017)

\bibitem{girshick2015fast}
Girshick, R.: Fast r-cnn. In: ICCV. pp. 1440--1448 (2015)

\bibitem{he2017mask}
He, K., Gkioxari, G., Doll{\'a}r, P., Girshick, R.: Mask r-cnn. In: ICCV. pp.
  2961--2969 (2017)

\bibitem{he2016deep}
He, K., Zhang, X., Ren, S., Sun, J.: Deep residual learning for image
  recognition. In: Proceedings of the IEEE conference on computer vision and
  pattern recognition. pp. 770--778 (2016)

\bibitem{ilg2017flownet}
Ilg, E., Mayer, N., Saikia, T., Keuper, M., Dosovitskiy, A., Brox, T.: Flownet
  2.0: Evolution of optical flow estimation with deep networks. In: CVPR. pp.
  2462--2470 (2017)

\bibitem{ilg2018occlusions}
Ilg, E., Saikia, T., Keuper, M., Brox, T.: Occlusions, motion and depth
  boundaries with a generic network for disparity, optical flow or scene flow
  estimation. In: ECCV. pp. 614--630 (2018)

\bibitem{insafutdinov2017arttrack}
Insafutdinov, E., Andriluka, M., Pishchulin, L., Tang, S., Levinkov, E.,
  Andres, B., Schiele, B.: Arttrack: Articulated multi-person tracking in the
  wild. In: CVPR. pp. 6457--6465 (2017)

\bibitem{kanade2000comprehensive}
Kanade, T., Cohn, J.F., Tian, Y.: Comprehensive database for facial expression
  analysis. In: FG. pp. 46--53. IEEE (2000)

\bibitem{li2014deepreid}
Li, W., Zhao, R., Xiao, T., Wang, X.: Deepreid: Deep filter pairing neural
  network for person re-identification. In: Proceedings of the IEEE conference
  on computer vision and pattern recognition. pp. 152--159 (2014)

\bibitem{lin2018prior}
Lin, C.C., Hung, Y.: A prior-less method for multi-face tracking in
  unconstrained videos. In: CVPR. pp. 538--547 (2018)

\bibitem{lin2017feature}
Lin, T.Y., Doll{\'a}r, P., Girshick, R., He, K., Hariharan, B., Belongie, S.:
  Feature pyramid networks for object detection. In: CVPR. pp. 2117--2125
  (2017)

\bibitem{lin2014microsoft}
Lin, T.Y., Maire, M., Belongie, S., Hays, J., Perona, P., Ramanan, D.,
  Doll{\'a}r, P., Zitnick, C.L.: Microsoft coco: Common objects in context. In:
  European conference on computer vision. pp. 740--755. Springer (2014)

\bibitem{liu2016ssd}
Liu, W., Anguelov, D., Erhan, D., Szegedy, C., Reed, S., Fu, C.Y., Berg, A.C.:
  Ssd: Single shot multibox detector. In: ECCV. pp. 21--37. Springer (2016)

\bibitem{ma2018trajectory}
Ma, C., Yang, C., Yang, F., Zhuang, Y., Zhang, Z., Jia, H., Xie, X.: Trajectory
  factory: Tracklet cleaving and re-connection by deep siamese bi-gru for
  multiple object tracking. In: ICME. pp.~1--6. IEEE (2018)

\bibitem{milan1603mot16}
Milan, A., Leal-Taix{\'e}, L., Reid, I., Roth, S., Schindler, K.: Mot16: A
  benchmark for multi-object tracking. arxiv 2016. arXiv preprint
  arXiv:1603.00831  \textbf{9}

\bibitem{milan2016mot16}
Milan, A., Leal-Taix{\'e}, L., Reid, I., Roth, S., Schindler, K.: Mot16: A
  benchmark for multi-object tracking. arXiv preprint arXiv:1603.00831  (2016)

\bibitem{ozerov2013evaluating}
Ozerov, A., Vigouroux, J.R., Chevallier, L., P{\'e}rez, P.: On evaluating face
  tracks in movies. In: ICIP. pp. 3003--3007. IEEE (2013)

\bibitem{pang2020tubetk}
Pang, B., Li, Y., Zhang, Y., Li, M., Lu, C.: Tubetk: Adopting tubes to track
  multi-object in a one-step training model. In: Proceedings of the IEEE/CVF
  Conference on Computer Vision and Pattern Recognition. pp. 6308--6318 (2020)

\bibitem{redmon2016you}
Redmon, J., Divvala, S., Girshick, R., Farhadi, A.: You only look once:
  Unified, real-time object detection. In: CVPR. pp. 779--788 (2016)

\bibitem{redmon2018yolov3}
Redmon, J., Farhadi, A.: Yolov3: An incremental improvement. arXiv preprint
  arXiv:1804.02767  (2018)

\bibitem{ren2015faster}
Ren, S., He, K., Girshick, R., Sun, J.: Faster r-cnn: Towards real-time object
  detection with region proposal networks. In: NIPS. pp. 91--99 (2015)

\bibitem{ristani2016performance}
Ristani, E., Solera, F., Zou, R., Cucchiara, R., Tomasi, C.: Performance
  measures and a data set for multi-target, multi-camera tracking. In: European
  Conference on Computer Vision. pp. 17--35. Springer (2016)

\bibitem{shen2017learning}
Shen, W., Liu, R.: Learning residual images for face attribute manipulation.
  In: CVPR. pp. 4030--4038 (2017)

\bibitem{sun2018pwc}
Sun, D., Yang, X., Liu, M.Y., Kautz, J.: Pwc-net: Cnns for optical flow using
  pyramid, warping, and cost volume. In: CVPR. pp. 8934--8943 (2018)

\bibitem{sun2019deep}
Sun, S., Akhtar, N., Song, H., Mian, A.S., Shah, M.: Deep affinity network for
  multiple object tracking. TPAMI  (2019)

\bibitem{tapaswi2019video}
Tapaswi, M., Law, M.T., Fidler, S.: Video face clustering with unknown number
  of clusters. In: ICCV. pp. 5027--5036 (2019)

\bibitem{voigtlaender2019mots}
Voigtlaender, P., Krause, M., Osep, A., Luiten, J., Sekar, B.B.G., Geiger, A.,
  Leibe, B.: Mots: Multi-object tracking and segmentation. In: CVPR. pp.
  7942--7951 (2019)

\bibitem{wang2019exploit}
Wang, G., Wang, Y., Zhang, H., Gu, R., Hwang, J.N.: Exploit the connectivity:
  Multi-object tracking with trackletnet. In: ACM MM. pp. 482--490. ACM (2019)

\bibitem{weng2020gnn3dmot}
Weng, X., Wang, Y., Man, Y., Kitani, K.M.: Gnn3dmot: Graph neural network for
  3d multi-object tracking with 2d-3d multi-feature learning. In: Proceedings
  of the IEEE/CVF Conference on Computer Vision and Pattern Recognition. pp.
  6499--6508 (2020)

\bibitem{Wojke2018deep}
Wojke, N., Bewley, A.: Deep cosine metric learning for person
  re-identification. In: WACV. pp. 748--756. IEEE (2018).
  \doi{10.1109/WACV.2018.00087}

\bibitem{wojke2017simple}
Wojke, N., Bewley, A., Paulus, D.: Simple online and realtime tracking with a
  deep association metric. In: ICIP. pp. 3645--3649. IEEE (2017)

\bibitem{wu2013simultaneous}
Wu, B., Lyu, S., Hu, B.G., Ji, Q.: Simultaneous clustering and tracklet linking
  for multi-face tracking in videos. In: ICCV. pp. 2856--2863 (2013)

\bibitem{xu2019deepmot}
Xu, Y., Ban, Y., Alameda-Pineda, X., Horaud, R.: Deepmot: A differentiable
  framework for training multiple object trackers. arXiv preprint
  arXiv:1906.06618  (2019)

\bibitem{yang2016wider}
Yang, S., Luo, P., Loy, C.C., Tang, X.: Wider face: A face detection benchmark.
  In: CVPR. pp. 5525--5533 (2016)

\bibitem{yin2020unified}
Yin, J., Wang, W., Meng, Q., Yang, R., Shen, J.: A unified object motion and
  affinity model for online multi-object tracking. In: Proceedings of the
  IEEE/CVF Conference on Computer Vision and Pattern Recognition. pp.
  6768--6777 (2020)

\bibitem{zhang2016tracking}
Zhang, S., Gong, Y., Huang, J.B., Lim, J., Wang, J., Ahuja, N., Yang, M.H.:
  Tracking persons-of-interest via adaptive discriminative features. In: ECCV.
  pp. 415--433. Springer (2016)

\bibitem{zhang2019robust}
Zhang, W., Zhou, H., Sun, S., Wang, Z., Shi, J., Loy, C.C.: Robust
  multi-modality multi-object tracking. In: Proceedings of the IEEE
  International Conference on Computer Vision. pp. 2365--2374 (2019)

\end{thebibliography}
}
\end{document}